\documentclass{article}

\PassOptionsToPackage{numbers, compress}{natbib}

\usepackage[preprint]{neurips_2019}

\usepackage[utf8]{inputenc} 
\usepackage[T1]{fontenc}    
\usepackage{hyperref}       
\usepackage{url}            
\usepackage{booktabs}       
\usepackage{amsfonts}       
\usepackage{nicefrac}       
\usepackage{microtype}      
\usepackage{graphicx}
\usepackage{amsmath}
\usepackage{amssymb}
\usepackage{mathtools}
\usepackage{commath}
\usepackage{algorithm}
\usepackage{multicol}
\usepackage[noend]{algpseudocode}

\usepackage{subcaption}\usepackage{multicol}
\usepackage{booktabs} 

\newtheorem{definition}{Definition}
\setcitestyle{square}

\title{Continual Learning via Online Leverage Score Sampling}

\author{%
  Dan Teng \\
  Neuri \\
  \texttt{dan@neuri.ai} \\
   \And
   Sakyasingha Dasgupta \\
   Neuri \\
   \texttt{sakya.dasgupta@gmail.com } \\
}

\begin{document}

\maketitle

\begin{abstract}
In order to mimic the human ability of continual acquisition and transfer of knowledge across various tasks, a learning system needs the capability for continual learning, effectively utilizing the previously acquired skills. As such, the key challenge is to transfer and generalize the knowledge learned from one task to other tasks, avoiding forgetting and interference of previous knowledge and improving the overall performance. In this paper, within the continual learning paradigm, we introduce a method that \textit{effectively forgets} the less useful data samples continuously and allows beneficial information to be kept for training of the subsequent tasks, in an online manner.
The method uses statistical leverage score information to measure the importance of the data samples in every task and adopts frequent directions approach to enable a continual or life-long learning property. This effectively maintains a constant training size across all tasks. We first provide mathematical intuition for the method and then demonstrate its effectiveness in avoiding catastrophic forgetting and computational efficiency on continual learning of classification tasks when compared with the existing state-of-the-art techniques. 
\end{abstract}

\section{Introduction}
\label{sec: intro}

It is a typical practice to design and optimize machine learning (ML) models to solve a single task. On the other hand, humans, instead of learning over isolated complex tasks, are capable of generalizing and transferring knowledge and skills learned from one task to another. This ability to remember, learn and transfer information across tasks is referred to as continual learning \cite{Thrun1995, Ruvolo2013, Hassabis2017, Parisi2019}. 
 The major challenge for creating ML models with continual learning ability is that they are prone to
\textit{catastrophic forgetting} \cite{McClelland1995, McCloskey1989, Goodfellow2013, French1999}. ML models tend to forget the knowledge learned from previous tasks when re-trained on new observations corresponding to a different (but related) task. Specifically when a deep neural network (DNN) is fed with a sequence of tasks, the ability to solve the first task will decline significantly after training on the following tasks. The typical structure of DNNs by design does not possess the capability of preserving previously learned knowledge without interference between tasks or catastrophic forgetting. In order to overcome catastrophic forgetting, a learning system is required to continuously acquire knowledge from the newly fed data as well as to prevent the training of the new data samples from destroying the existing knowledge.

In this paper, we propose a novel approach to continual learning with DNNs that addresses the catastrophic forgetting issue, namely a technique called \textit{online leverage score sampling (OLSS)}. In OLSS, we progressively compress the input information learned thus far, along with the input from current task and form more efficiently condensed data samples. The compression technique is based on the statistical leverage scores measure, and it uses the concept of frequent directions in order to connect the series of compression steps for a sequence of tasks. 

When thinking about continual learning, a major source of inspiration is the ability of biological brains to learn without destructive interference between older memories and generalize knowledge across multiple tasks. In this regard, the typical approach is enabling some form of episodic-memory in the network and consolidation \cite{McClelland1995} via replay of older training data. However, this is an expensive process and does not scale well for learning large number of tasks. As an alternative, taking inspiration from the neuro-computational models of complex synapses \cite{BennaFusi2016}, recent work has focused on assigning some form of importance to parameters in a DNN and perform task-specific synaptic consolidation \cite{Kirkpatrick2017, Zenke2017}. Here, we take a very different view of continual learning and find inspiration in the brains ability for dimensionality reduction \cite{Pang2016} to extract meaningful information from its environment and drive behavior. As such, we enable such progressive dimensionality reduction (in terms of number of samples) of previous task data combined with new task data in order to only preserve a good summary information (discarding the less relevant information or effective forgetting) before further learning. Repeating this process in an online manner we enable continual learning for a large sequence of tasks. Much like our brains, a central strategy employed by our method is to strike a balance between dimensionality reduction of task specific data and dimensionality expansion as processing progresses throughout the hierarchy of the neural network \cite{FusiMiller2016}. 

\subsection{Related Work}
Recently, a number of approaches have been proposed to adapt a DNN model to the continual learning setting, from an adaptive model architecture perspective such as adding columns or neurons for new tasks \cite{Rusu2016, Yoon2018, Schwarz2018}; model parameter adjustment or regularization techniques like, imposing restrictions on parameter updates \cite{Kirkpatrick2017, Zenke2017, Li2016, Titsias2019}; memory revisit techniques which ensure model updates towards the optimal directions \cite{Lopez-Paz2017, Rebuffi2017, Shin2017}; Bayesian approaches to model continuously acquired information \cite{Titsias2019, Nguyen2018, Garnelo2018}; or on broader domains with approaches targeted at different setups or goals such as few-shot learning or transfer learning \cite{Finn2017, Nichol2018}. 

In order to demonstrate our idea in comparison with the state-of-the-art techniques, we briefly discuss the following three popular approaches to continual learning: \\
I) \textbf{Regularization}: It constrains or regularizes the model parameters by adding additional terms in the loss function that prevent the model from deviating significantly from the parameters important to earlier tasks. Typical algorithms include elastic weight consolidation (EWC) \cite{Kirkpatrick2017} and continual learning through synaptic intelligence (SI) \cite{Zenke2017}. \\
II) \textbf{Architectural modification}: It revises the model structure successively after each task in order to provide more memory and additional free parameters in the model for new task input. Recent examples in this direction are progressive neural networks \cite{Rusu2016} and dynamically expanding networks \cite{Yoon2018}.\\
III) \textbf{Memory replay}: It stores data samples from previous tasks in a separate memory buffer and retrains the new model based on both the new task input and the memory buffer. 
Popular algorithms here are gradient episodic memory (GEM) \cite{Lopez-Paz2017}, incremental classifier and representation learning (iCaRL) \cite{Rebuffi2017}. 

Among these approaches, regularization is particularly prone to saturation of learning when the number of tasks is large.  The additional / regularization term in the loss function will soon lose its competency when important parameters from different tasks are overlapped too many times. Modifications on network architectures like progressive networks resolve the saturation issue, but do not scale when the number and complexity of tasks increase. The scalability problem is also prominent when using current memory replay techniques, often suffering from high memory and computational costs.

Our approach resembles the use of memory replay since it preserves the original input data samples from earlier tasks for further training. However, it does not require extra memory for training and is cost efficient compared to previous memory replay methods. It also makes more effective use of the model structure by exploiting the model capacity to adapt with more tasks, in contrast to constant addition of neurons or additional network layers for new tasks. Furthermore, unlike the importance assigned to model specific parameters when using regularization methods, we assign importance to the training data that is relevant in effectively learning new tasks, while forgetting less important information.  

\section{Online Leverage Score Sampling}
\label{sec: olss}
Before presenting the idea, we first setup the problem:
Let $\{(A_1, B_1), (A_2, B_2), ..., (A_i, B_i), ...\}$ represent a sequence of tasks, each task consists of $n_i$ data samples and each sample has a feature dimension $d$ and an output dimension $m$, i.e., input $A_i\in\mathbb{R}^{n_i\times d}$ and true output $B_i\in\mathbb{R}^{n_i\times m}$. Here, we assume the feature and output dimensions are fixed for all tasks \footnote{If we know apriori that the feature or output dimensions are different, we could choose a presumed larger value of $d$ and $m$. In continuous learning our aim is to solve successive problems with some degree of overlap. As such, the feature and output dimensions being the same across tasks is not overly strict.}. 
The goal is to train a DNN over the sequence of tasks and ensure it performs well on all of them, without catastrophic forgetting. Here, we consider that the network's architecture stays the same and the tasks are received in a sequential manner.
Formally, with $f$ representing a DNN, our objective is to minimize the loss \footnote{Here, we represent a generic Euclidean loss term. However, this could take the form of any typical formulation in terms of $l_1$-loss, $l_2$-loss or cross-entropy loss as commonly used in classification problems.}: 
\begin{equation}
\label{eq: main}
\min_f\norm{f(A) - B}_2^2 \text{ where } A = \begin{bmatrix} A_1 \\ A_2 \\... \\ A_i \\ ...
\end{bmatrix} \text{ and } B = \begin{bmatrix} B_1 \\ B_2 \\ ... \\ B_i \\ ... \end{bmatrix}.
\end{equation} 
Under this setup, we look at some of the existing models:

Online EWC trains $f$ on the $i$th task $(A_i, B_i)$ with a loss function containing additional penalty terms 
\[
\min_f\norm{f(A_i) - B_i}_2^2 + \sum_{j=1}^{i-1}\sum_{p=1}^{w} \lambda F_p^j (\theta_p - \theta_p^{j*})^2, 
\]
where $\lambda$ indicates the importance level of the previous tasks compared to task $i$, $F_p^j$ represents the $p$th diagonal entry of the Fisher information matrix for Task $j$, $w$ represents the number of parameters in the network, $\theta_p$ corresponds to the $p$th model parameter for the current task and $\theta_p^{j*}$ is the $p$th model parameter value for the $j$th task. 

Alternately, GEM maintains an extra memory buffer containing data samples from each of the previous tasks $\mathcal{M}_k$ with $k < i$. It trains on the current task $(A_i, B_i)$ with a regular loss function, but subject to inequalities on each update of $f$ (update on each parameter $\theta$),
\begin{eqnarray*}
    && \min_f \norm{f(A_i) - B_i}_2^2 \\
    \text{s. t.} && \Bigg\langle\dfrac{\partial\ \norm{f_\theta(A_i) - B_i}_2^2}{\partial\theta}, \dfrac{\partial\  \norm{f_\theta(A_{\mathcal{M}_k}) - B_{\mathcal{M}_k}}_2^2}{\partial\theta}\Bigg\rangle \geq 0 \text{ for all } k<i.
\end{eqnarray*}

\subsection{Our approach}
The new method OLSS, different from either method above, 
targets to find an approximation of $A$ in a streaming (online) manner, i.e., form a sketch $\hat{A}_i\in\mathbb{R}^{\ell\times d}$ to approximate $[A_1^T ~ A_2^T ~~ \cdots ~~ A_i^T]^T\in\mathbb{R}^{(n_1+...+n_i)\times d}$ such that the resulting 
\[\hat{f}_i := \arg\min_f\norm{f(\hat{A}_i) - \hat{B}_i}_2^2
\]
is likely to perform on all tasks as good as 
\begin{equation}
f^*_i := \arg\min_f\norm{f([A_1^T ~ A_2^T ~~ \cdots ~~ A_i^T]^T) - [B_1^T ~ B_2^T ~~ \cdots ~~ B_i^T]^T}_2^2.
\label{eq: fi}
\end{equation}
In order to avoid extra memory and computation cost during the training process, we could set the approximate $\hat{A_i}$ to have the same number of rows (number of data samples) as the current task $A_i$. 

Equation (\ref{eq: main}) and (\ref{eq: fi}) represent nonlinear least squares problems. It is to be noted that a nonlinear least squares problem can be solved with an approximation deduced from an iteration of linear least squares problems with $J^TJ \Delta \theta = J^T\Delta B$ where $J$ is the Jacobian of $f$ at each update (using the Gauss-Newton method). Besides this technique, there are various other approaches in addressing this problem. Here we adopt a cost effective simple randomization technique - leverage score sampling, which has been used extensively in solving large scale linear least squares and low rank approximation problems \cite{Cohen2017, Drineas2012, Woodruff2014}. 


\subsection{Statistical Leverage Score and Leverage Score Sampling}
\begin{definition}\cite{Drineas2012}
Given a matrix $A\in \mathbb{R}^{n\times d}$ with $n> d$, let $U$ denote the $n\times d$ matrix consisting of the $d$ left singular vectors of $A$, and let $U_{(i,:)}$ denote the $i$-th row of $U$, then the statistical leverage score of the $i$-th row of $A$ is defined as $\norm{U_{(i,:)}}_2^2$ for $i \in \{1, ..., n\}$. 
\end{definition}
Statistical leverage scores measure the non-uniformity structure of a matrix and a higher score indicates a heavier weight of the row contributing to the non-uniformity of the matrix. It has been widely used for outlier detection in statistical data analysis. In recent applications \cite{Drineas2012, Woodruff2014}, it also emerges as a fundamental tool for constructing randomized matrix sketches. Given a matrix $A\in\mathbb{R}^{n\times d}$, a sketch of $A$ is another matrix $B\in\mathbb{R}^{\ell\times d}$ where $\ell$ is significantly smaller than $n$ but still approximates $A$ well, more specifically, $\norm{A^TA - B^TB}_2 \leq \varepsilon \norm{A}_2^2$. Theoretical accuracy guarantees have been derived for random sampling methods based on statistical leverage scores \cite{Woodruff2014, Ma2014}. 

Considering our setup which is to approximate a matrix for solving a least squares problem and also the computational efficiency, we adopt the following leverage score based sampling method: \\
Given a sketch size $\ell$, define a distribution $\{p_i, ..., p_n\}$ \footnote{Since $d = \norm{U}_F^2 = \sum_{i=1}^n \norm{U_{(i,:)}}_2^2$, $\{p_1, ..., p_n\}$ forms a probability distribution.} with $p_i = \frac{\norm{U_{(i,:)}}_2^2}{d}$, 
 the sketch is formed by
 independently and randomly selecting $\ell$ rows of $A$ without replacement, where the $i$th row is selected with probability $p_i$. Based on this, we are able to select the samples that contributes the most to a given dataset. The remaining problem is to embed it in a sequence of tasks and still generate promising approximations to solve the least squares problem. In order to achieve that, we make use of the concept of frequent directions. 

\subsection{Frequent Directions}
Frequent directions extends the idea of frequent items in item frequency approximation problem to a matrix \cite{Liberty2013, Ghashami2016, Teng2019} and it is also used to generate a sketch for a matrix, but in a data streaming environment. As the rows of $A\in\mathbb{R}^{n\times d}$ are fed in one by one, the original idea of frequent directions is to first perform Singular Value Decomposition (SVD) on the first $2\ell$ rows of $A$ and shrink the top $\ell$ singular values by the same amount which is determined by the $(\ell+1)$th singular value, and then save the product of the shrunken top $\ell$ singular values and the top  $\ell$ right singular vectors as a sketch for the first $2\ell$ rows of $A$. With the next $\ell$ rows fed in, append them behind the sketch and perform the shrink and product. This process is repeated until reaching the final sketch $\hat{A}\in\mathbb{R}^{\ell\times d}$ for $A\in\mathbb{R}^{n\times d}$. Different from the leverage score sampling sketching technique, a deterministic bound is guaranteed for the accuracy of the sketch: $\norm{A^TA - \hat{A}^T\hat{A}}_2^2\leq\norm{A-A_k}_F^2/(\ell-k)$ with $l>k$ and $A_k$ denotes best rank-$k$ approximation of $A$ \cite{Liberty2013, Ghashami2016}.

Inspired by the routine of frequent directions in a streaming data environment, our OLSS method is constructed as follows: First initialize a `sketch' matrix $\hat{A}\in\mathbb{R}^{\ell\times d}$ and a corresponding $\hat{B}\in\mathbb{R}^{\ell\times m}$. For the first task $(A_1\in\mathbb{R}^{n_1\times d}, B_1\in\mathbb{R}^{n_1\times m})$, we randomly select $\ell$ rows of $A$ and (the corresponding $\ell$ rows of) $B$ without replacement according to the leverage score sampling defined above with probability distribution based on $A$'s leverage scores, then train the model on the sketch ($\hat{A}$, $\hat{B}$); after seeing Task 2, we append ($A_2$, $B_2$) to the sketch ($\hat{A}$, $\hat{B}$) respectively and again randomly select $\ell$ out of $\ell + n_2$ data samples according to the leverage score sampling with the probability distribution based on the leverage scores of $[\hat{A}^T, A_2^T]^T \in\mathbb{R}^{(\ell+n_2)\times d}$, and form a new sketch $\hat{A}\in\mathbb{R}^{\ell\times d}$ and $\hat{B}\in\mathbb{R}^{\ell\times d} $, then train on it. This process is repeated until the end of the task sequence. We present the step by step procedure in Algorithm \ref{alg: olss}.

\subsection{Main Algorithm}
\label{sec: main}
The original idea of leverage score sampling and frequent directions both have the theoretical accuracy bounds with the sketch on the error term $\norm{A^TA - \hat{A}^T\hat{A}}_2$. The bounds show that the sketch $\hat{A}$ contains the relevant information used to form the covariance matrix of all the data samples $A^TA$, in other words, the sketch captures the relationship among the data samples in the feature space (which is of dimension $d$). 
For a sequence of tasks, it is common to have noisy data samples or interruptions among samples for different tasks. The continuous update of important rows in a matrix (data samples for a sequence of tasks), or the continuous effective forgetting of less useful rows may serve as a filter to remove the unwanted noise. 

Different from most existing methods, Algorithm \ref{alg: olss} does not work directly with the training model, instead it could be considered as data pre-processing which constantly extracts useful information from previous and current tasks. Because of its parallel construction, OLSS could be combined with all the aforementioned algorithms to further improve its performance.




\begin{algorithm}
\small
\caption{Online Leverage Score Sampling}
\label{alg: olss}
\begin{algorithmic}[1]
\Require A sequence of tasks $\{(A_1, B_1), ..., (A_i, B_i), ...\}$ with $A_i\in\mathbb{R}^{n_i\times d}$ and $B_i\in\mathbb{R}^{n_i\times m}$; initialization of the model parameters; a space parameter $\ell$ i.e., number of samples to pass in the model for training. It can be set as $n_i$ or even smaller after receiving the $i$-th task, which avoids extra memory and computations during training.
\Ensure A trained neural network on a sequence of tasks.
\State Initialize a buffer set $S= \{\hat{A}, \hat{B}\}$ where both $\hat{A}$ and $\hat{B}$ are empty.
\While {the $i$th task is presented}
\If {$\hat{A}$ and $\hat{B}$ are empty} 
\State set $\hat{A} = A_i$ and $\hat{B}= B_i$,
\Else 
\State set $\hat{A} = \begin{bmatrix} \hat{A} \\ A_i\end{bmatrix}$ and $\hat{B} = \begin{bmatrix} \hat{B} \\ B_i\end{bmatrix}$.
\EndIf
\State Perform SVD: $[U, \Sigma, V^T] = svd(\hat{A})$.
\State Randomly select $\ell$ rows of $\hat{A}$ and $\hat{B}$ without replacement based on probability  $\norm{U_{j,:}}_2^2/\norm{U}_F^2$ for $j\in \{1, ..., n_i+\ell\}$ (or $j\in\{1, ..., n_i\}$ when $i=1$) and set them as $\hat{A}$ and $\hat{B}$ respectively.
\State Train the model with $\hat{A}\in\mathbb{R}^{\ell\times d}$ and $\hat{B}\in\mathbb{R}^{\ell\times m}$.
\EndWhile
\end{algorithmic}
\end{algorithm}

Regarding the computational complexity, when $n_i$ is large, the SVD of $\hat{A}\in\mathbb{R}^{(n_i+\ell)\times d}$ in Step 6 is computationally expensive which takes $O((n_i+\ell)d^2)$ time. This procedure is for the computation of leverage scores which can be sped up significantly with various leverage score approximation techniques in the literature \cite{Drineas2012, Cohen2017, Rudi2018}, such as through the randomized algorithm in \cite{Drineas2012}, the leverage scores for $\hat{A}$ could be approximated in $O((n_i+\ell)d\log(n_i+\ell))$ time. 




 However, one possible drawback of the above procedure is that the relationship represented in a covariance matrix is linear, so any underlying nonlinear connections among the data samples may not be fully captured in the sketch. Furthermore, the structure of the function $f$ would also affect the information required to be kept in the sketch in order to perform well on solving the least squares problem in (\ref{eq: fi}). As such, there may exist certain underlying dependency of a data sample's importance on the DNN model architecture. This remains a future research direction. 
 


\section{Experiments}
\label{sec: exp}
We evaluate the performance of the proposed algorithm OLSS on three classification tasks used as benchmarks in related prior work.
\begin{itemize}
\itemsep0em 
    \item \textbf{Rotated MNIST} \cite{Lopez-Paz2017}: a variant of the MNIST dataset of handwriten digits \cite{LeCun1998}, the digits in each task are rotated by a fixed angle between $0^\circ$ to $180^\circ$. The experiment is on $20$ tasks and each task consists of $60,000$ training and $10,000$ testing samples.  
    \item \textbf{Permutated MNIST} \cite{Kirkpatrick2017}: a variant of the MNIST dataset \cite{LeCun1998}, the digits in each task are transformed by a fixed permutation of pixels. The experiment is on $20$ tasks and each task consists of $60,000$ training and $10,000$ testing samples.  
    \item \textbf{Incremental CIFAR100} \cite{Rebuffi2017, Zenke2017}: a variant of the CIFAR object recognition dataset with $100$ classes \cite{Krizhevsky2009}. The experiment is on $20$ tasks and each task consists of $5$ classes; each task consists of $2,500$ training and $500$ testing samples. Where, each task introduces a new set of classes; for a total number of 20 tasks, each new task concerns examples from a disjoint subset of 5 classes.
\end{itemize}
In the setting of \cite{Lopez-Paz2017} for incremental CIFAR100, a softmax layer is added to the output vector which only allows entries representing the $5$ classes in the current task to output values larger than $0$. In our setting, we allow the entries representing all the past occurring classes to output values larger than $0$. We believe this is a more natural setup for continual learning. 

For the aforementioned experiments, we compare the performance of the following algorithms: 
\begin{itemize}
    \item A simple SGD predictor.
    \item EWC \cite{Kirkpatrick2017}, as discussed earlier in Section \ref{sec: olss}.
    \item GEM \cite{Lopez-Paz2017}, as discussed earlier in Section \ref{sec: olss}.
    \item iCaRL \cite{Rebuffi2017}, it classifies based on a nearest-mean-of-exemplars rule, keeps an episodic memory and updates its exemplar set continuously to prevent catastrophic forgetting. It is only applicable to incremental CIFAR100 experiment due to its requirement on the same input representation across tasks.
    \item OLSS (ours).
\end{itemize}

In addition to these, experiments were also conducted using SI \cite{Zenke2017} and the same three tasks. However, no significant improvement in performance and a sensitivity to learning rate parameter choice was observed, with learning ability being relatively better than online EWC. As such we don’t show SI performance in our plots. It can however be tested using our open sourced code\footnote{The PyTorch code base implementing all the experiments presented here will be made publicly available on Github.} for this paper. 

The competing algorithms SGD, EWC, GEM and iCaRL were implemented based on the publicly available code from the original authors of the GEM paper \cite{Lopez-Paz2017}; a plain SGD optimizer is used for all algorithms. The DNN used for rotated and permuted MNIST is an MLP with $2$ hidden layers and each with $400$ rectified linear units; whereas a smaller version of ResNet18 \cite{Heetal2016}, with three times less feature maps across all layers is used for the incremental CIFAR100 experiment. We train $5$ epochs with batch size $200$ on rotated and permuted MNIST datasets and $10$ epochs with batch size $100$ on incremental CIFAR100.  The regularization and memory hyper-parameters in EWC, iCaRL and GEM were set as described in \cite{Lopez-Paz2017}. The space parameter for our OLSS algorithm was set to be equal to the number of samples in each task. The learning rate for each algorithm was determined through a grid search on $\{0.001, 0.003, 0.01, 0.03, 0.1, 0.3, 1.0\}$. The final learning rates used in each experiment corresponding to the different algorithms was set as, \\
rotated MNIST, SGD: $0.1$, EWC: $0.3$, GEM: $0.3$ and OLSS: $0.3$; permutated MNIST, SGD: $0.001$, EWC: $0.1$, GEM: $0.1$ and OLSS: $0.1$ and for incremental CIFAR100, SGD: $0.1$, EWC: $0.1$, GEM: $0.3$, iCaRL: $0.01$ and OLSS: $1.0$.

\subsection{Results}
\begin{figure}[h]
    \centering
    \includegraphics[width=6.9cm]{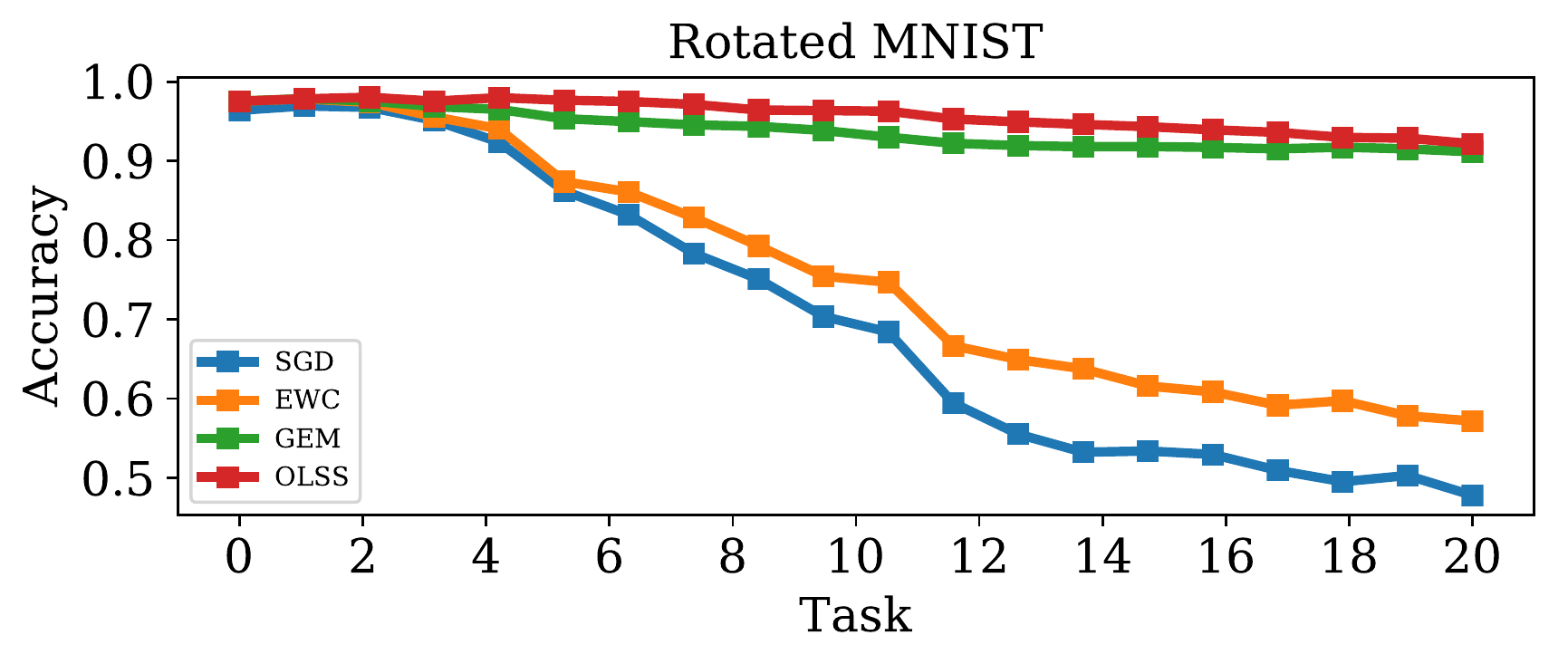}
    \includegraphics[width=6.9cm]{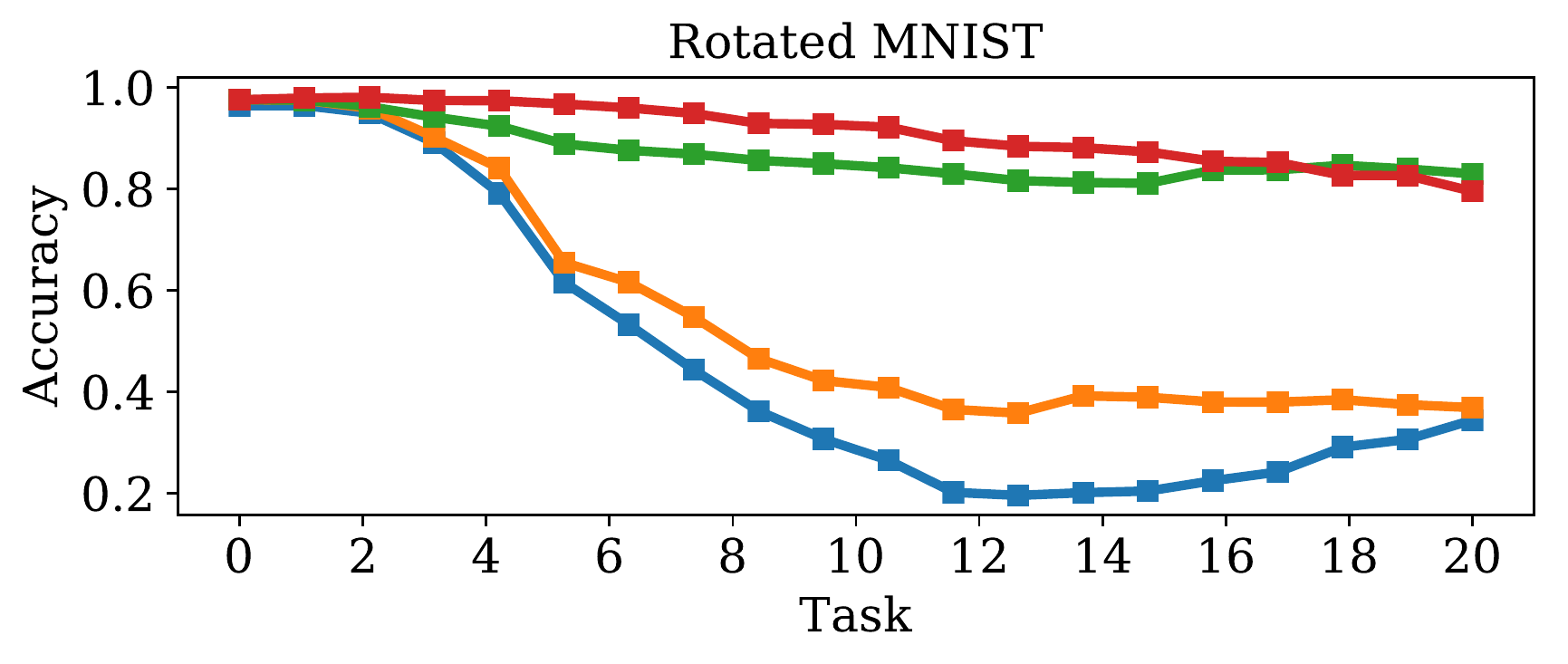}
    \includegraphics[width=6.9cm]{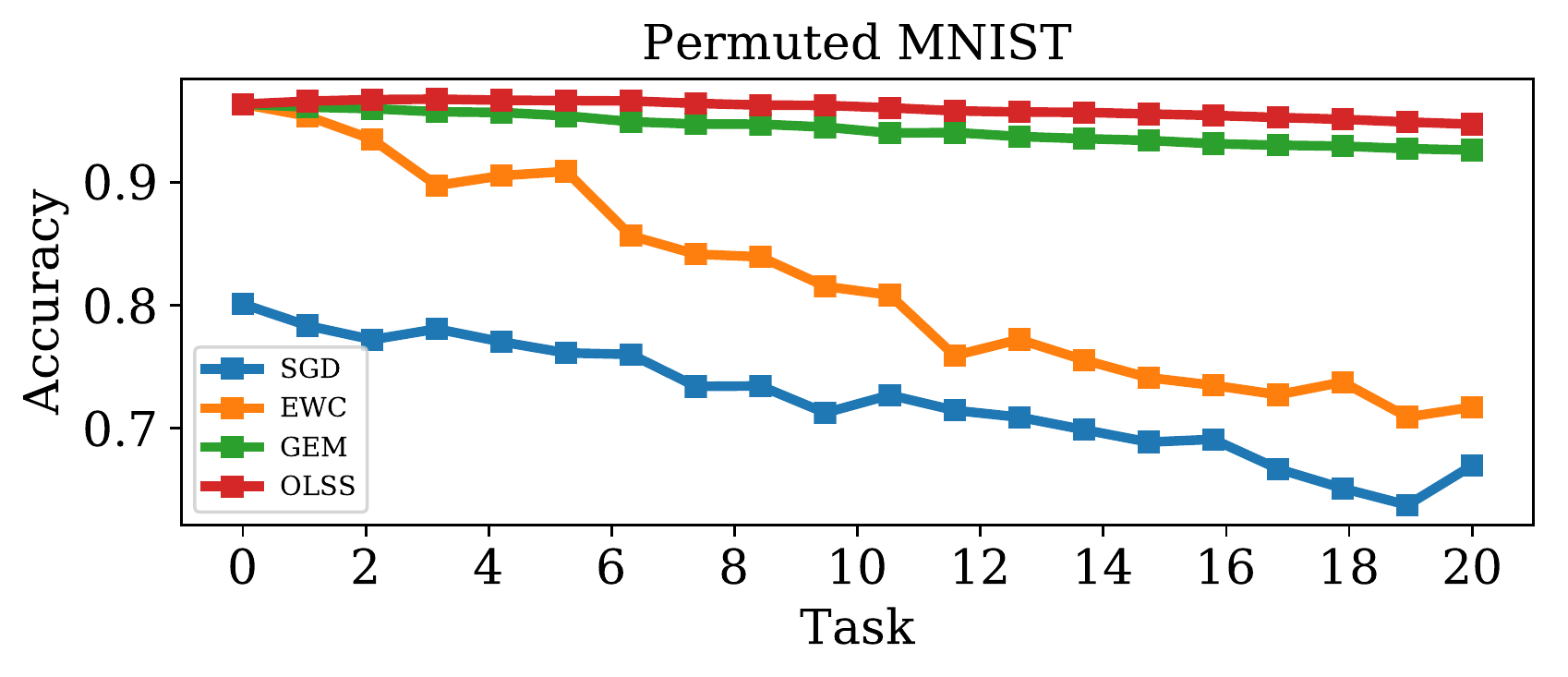}
    \includegraphics[width=6.9cm]{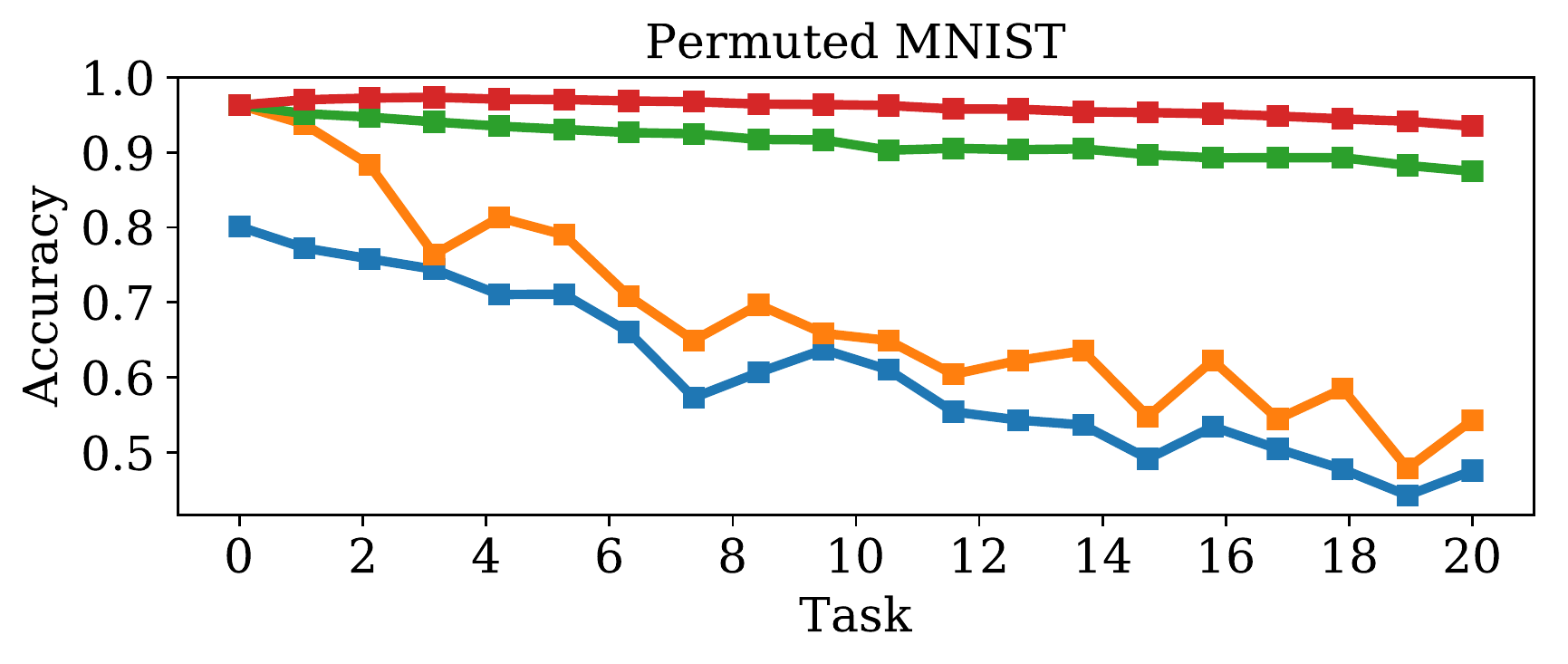}
    \includegraphics[width=6.9cm]{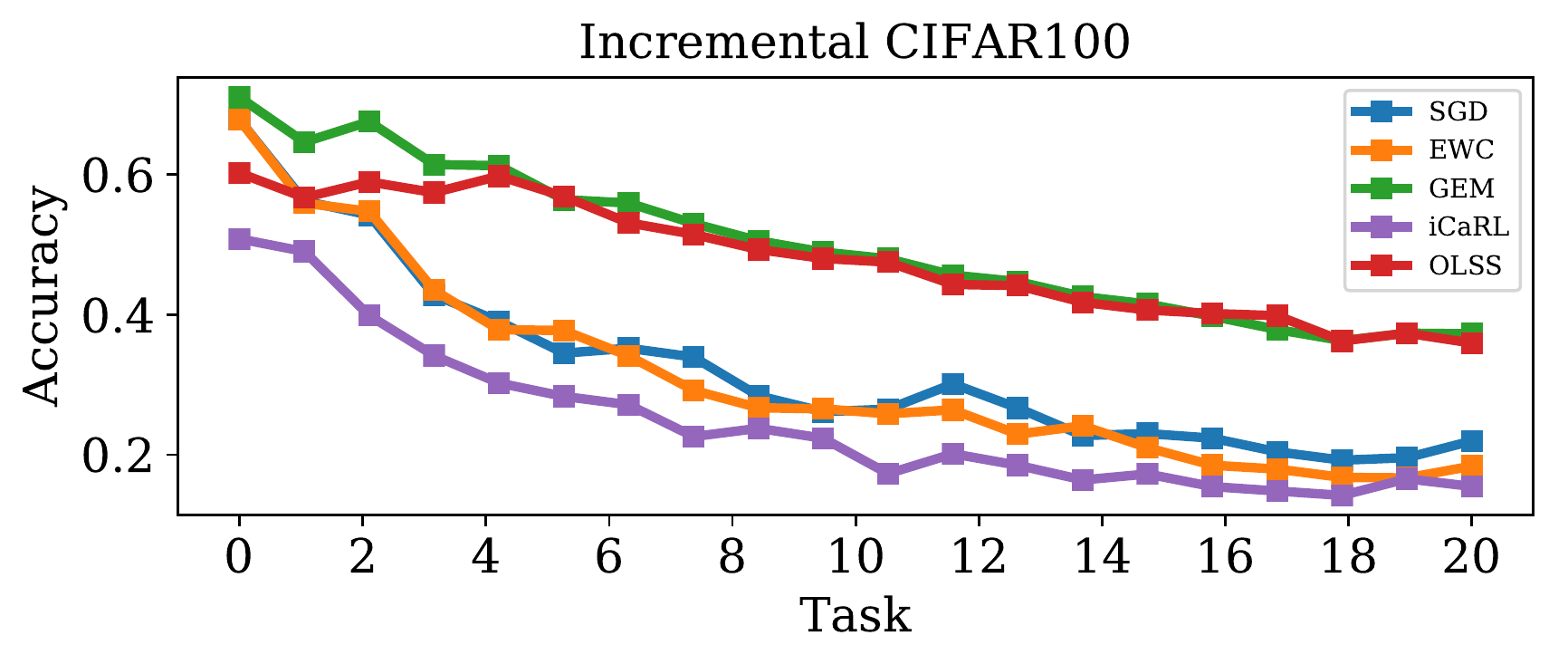}
    \includegraphics[width=6.9cm]{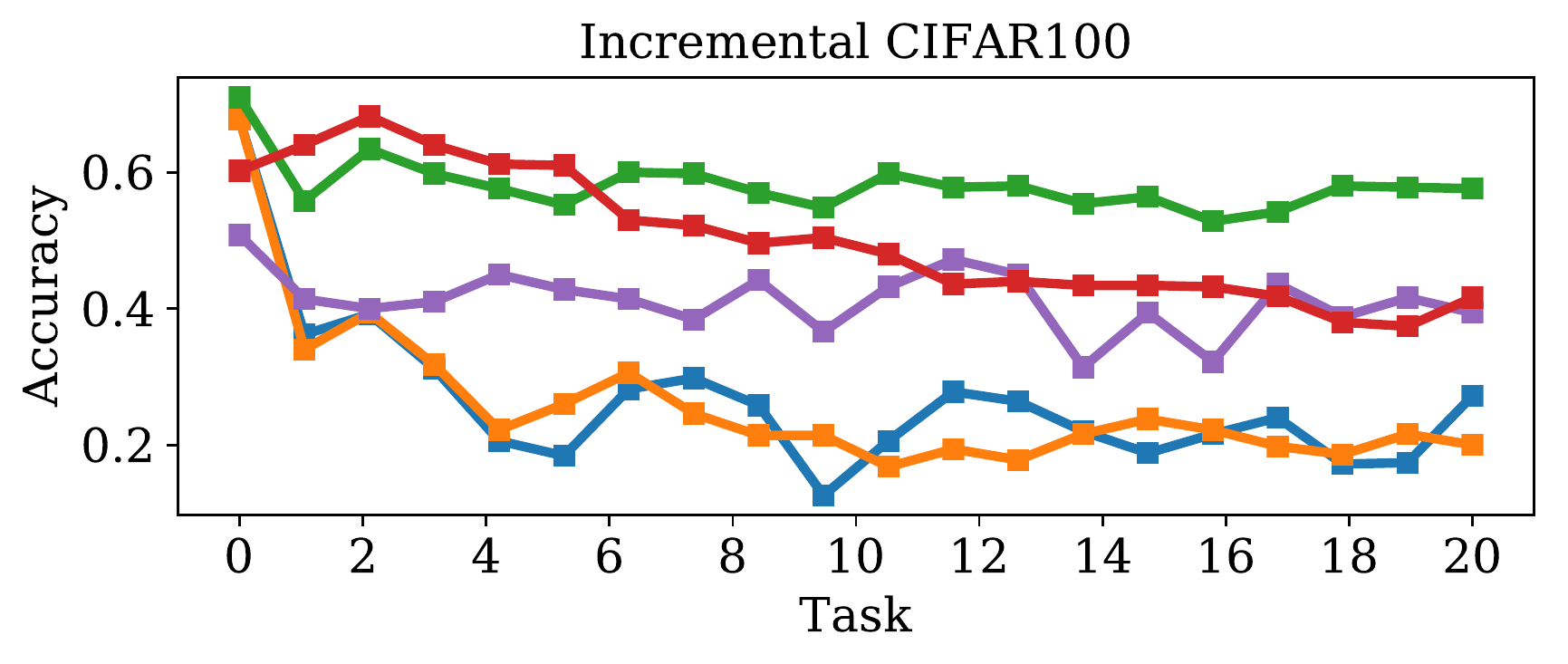}
    \caption{\textbf{Left}: Evolution of average test accuracy across all the learned tasks after training on a sequence of tasks. (E.g., the accuracy value at Task $=10$ is the average test accuracy on Task $1-10$ after training the model for $10$ consecutive tasks.)
    \textbf{Right}: Evolution of test accuracy for the first task after training on a sequence of tasks. (E.g., the accuracy value at Task $=10$ is the accuracy of Task $1$ after training the model for $10$ consecutive tasks.)}%
    \label{fig: task_avg}
\end{figure}

To evaluate the performance of different algorithms, we examine
\begin{itemize}
    \item The average test accuracy (Figure \ref{fig: task_avg} (left)), defined as $\frac{1}{k}\sum_{i=1}^k Acc(\text{task } i)$ after training $x=k$ tasks.
    \item Task 1's test accuracy (Figure \ref{fig: task_avg} (right)), defined as $Acc(\text{task } 1)$ after training $x=k$ tasks.
    \item Wall clock time (Table \ref{tab: time}).
\end{itemize}


As observed from Figure \ref{fig: task_avg} (left) across the three benchmarks, OLSS achieves similar average task accuracy or slightly higher compared to GEM and clearly outperforms SGD, EWC and iCaRL. This demonstrates the the ability of OLSS for continuously selecting useful data samples with progressive learning to overcome the catastrophic forgetting issue. In terms of maintaining the performance of the earliest task (Task 1) after training a sequence of tasks, OLSS shows the most robust performance at par with GEM on rotated and permutated MNIST, and slightly worse than GEM as the number of tasks increases in case of incremental CIFAR100. However, both these methods, significantly outperform SGD, EWC and iCaRL. 

In order to compare the computational time complexity across the methods, we report the walk clock time in Table \ref{tab: time}. Noticeably, SGD is the fastest among all the algorithms, however performs the worst as observed in Figure \ref{fig: task_avg}, then followed by OLSS and EWC (only in the case of CIFAR100, EWC is relatively faster than OLSS). The algorithms iCaRL and GEM both demand much higher computational costs, with GEM being significantly slow compared to the rest. This behavior is expected due to the requirement of additional constraint validation and at certain occasions, a gradient projection step (in order to correct for constraint violations across data samples from previously learned tasks stored in the memory buffer) in GEM (see Section 3 in \cite{Lopez-Paz2017}). As such although the buffered replay-memory based approach in GEM prevents catastrophic forgetting, the computational cost becomes prohibitively slow to be performed online while training DNNs on sequential multi-task learning scenarios.

Based on the performance and computational efficiency on all three datasets, OLSS emerges as the most favorable among the current state of the art algorithms for continual learning.  

\begin{table}[h]
\tiny
\caption{Wall Clock Time (s)}
\label{tab: time}
\begin{center}
\begin{small}
\begin{sc}
\begin{tabular}{lcccr}
\toprule
    & Rotated & Permuted  & Incremental \\
    &  MNIST &  MNIST & CIFAR100 \\
\midrule
SGD    & $158$& $152$& $780$ \\
EWC     & $944$& $896$ & $1213$ \\
GEM    & $8688$& $8846$ & $17868$ \\
ICARL  & -  &  - & $2802$ \\
OLSS     & $496$ & $455$ & $1363$       \\
\bottomrule
\end{tabular}
\end{sc}
\end{small}
\end{center}
\vskip -0.1in
\end{table}

\subsection{Discussions}
The space parameter of OLSS ($\ell$ in Algorithm \ref{alg: olss}) could be varied to balance its accuracy and efficiency. Here the choice of $\ell=n_i$ (number of samples in current task) is selected such that the number of training samples would be standardized across all algorithms, enabling effective compression and extraction of data samples for OLSS in a straightforward comparison. However, it is to be noted that if $\ell=n_i$, OLSS indeed requires some additional memory in order to compute the SVD of concatenated sketch of previous tasks and the current task. Unless, the algorithm is run in an edge computing environment with limited memory on chip, this issue could be ignored. 

On the other hand, GEM and iCaRL keep an extra episodic memory throughout the training process. Memory size was set to be $256$ for GEM and $1280$ for iCaRL by considering the accuracy and efficiency in the experiments. Variations on the size of the episodic memory would also affect their performance as well as the running time.  
As described earlier, GEM requires a constraint validation step and a potential gradient projection step for every update of the model parameters. As such the computational time complexity in this case is proportional to the product of the number of samples kept in the episodic memory, the number of parameters in the model and the number of iterations required to converge. In contrast, OLSS uses a SVD to compute the leverage scores for each task which can be achieved in a time complexity proportional to the product of the square of the number of features and the number of data samples. This is considerably less compared to GEM as shown in Table \ref{tab: time}. The computational complexity can be further reduced with fast leverage score approximation methods like randomized algorithm in \cite{Drineas2012}. 

\begin{figure}[h]
    \centering
    \includegraphics[width=6.6cm]{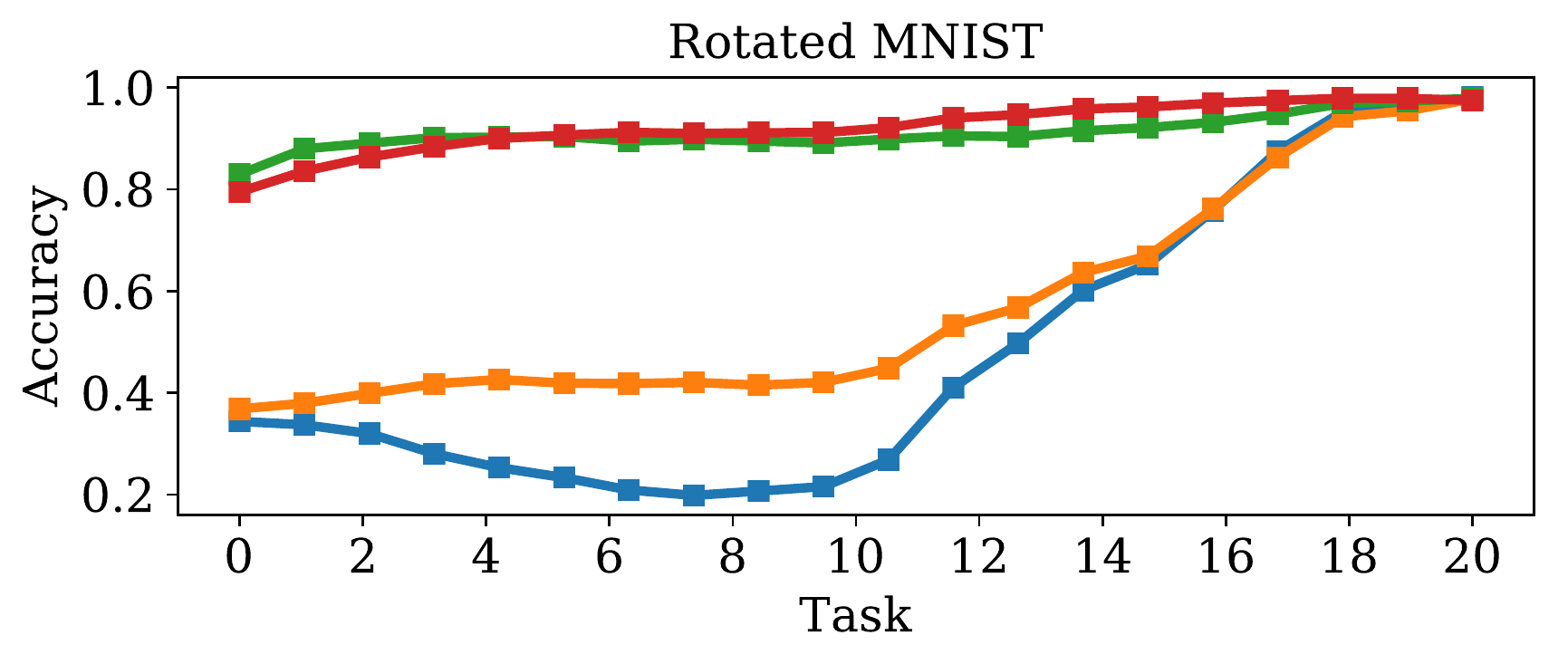}
    \includegraphics[width=6.6cm]{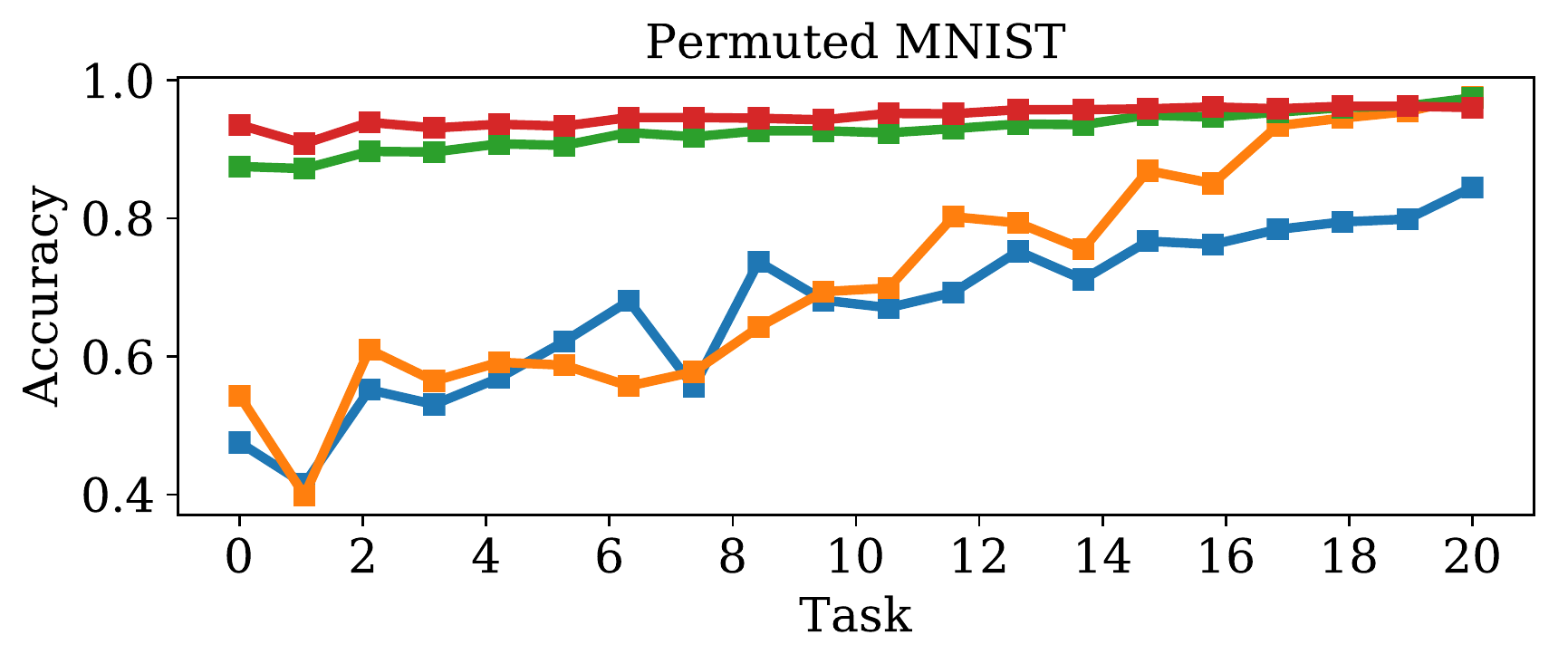}
    \includegraphics[width=6.6cm]{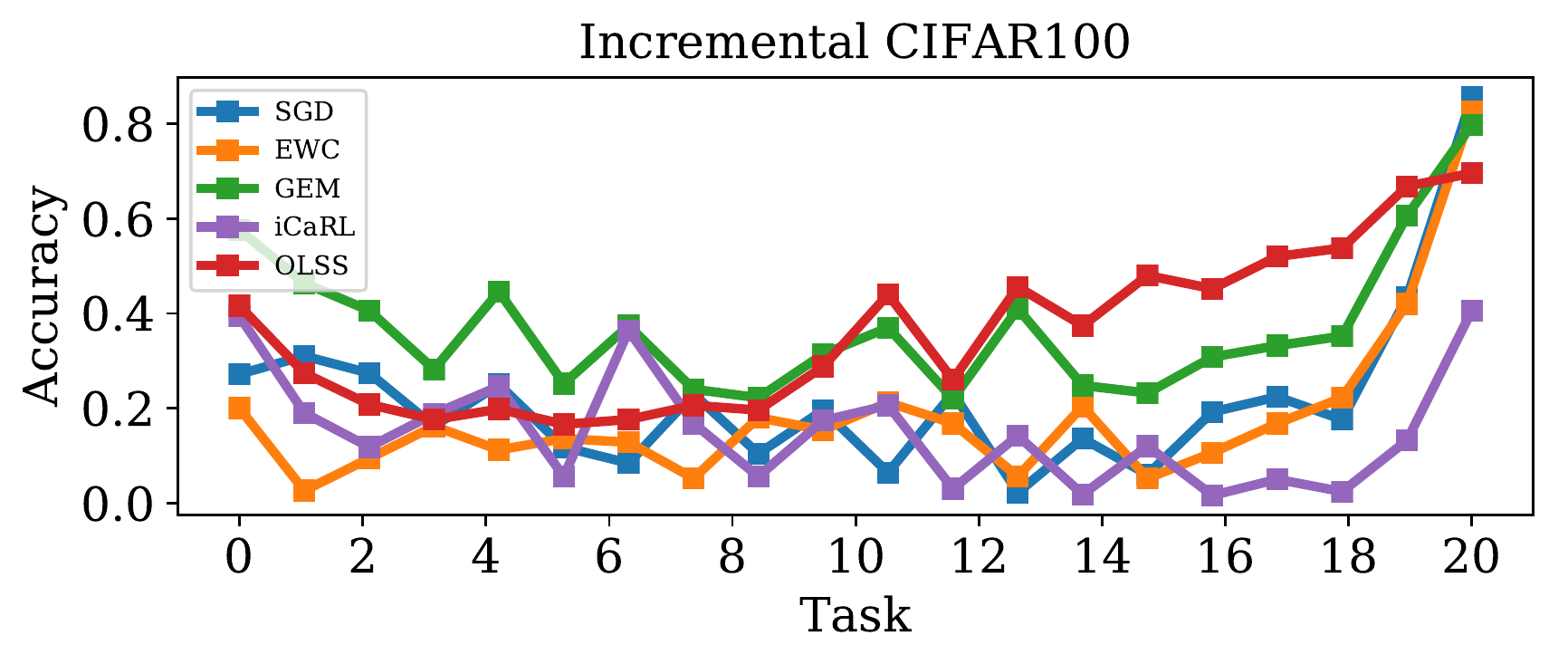}
    \caption{Accuracy of each task after training sequentially on all tasks. (E.g., the accuracy value at Task = 10 is the accuracy of Task 10 after training the model on all tasks sequentially.)}
    \label{fig: satur}
\end{figure}

As shown in Figure \ref{fig: satur}, after training the whole sequence of tasks, both GEM and OLSS are able to preserve the accuracy for most tasks on rotated and permuted MNIST. Nevertheless, it is difficult to completely recover the accuracy of previously trained tasks on CIFAR100 for all algorithms. In case of synaptic consolidation based method like EWC, the loss function contains additional regularization or penalty terms for each previously trained tasks. These additional penalties are isolated from each other. As the number of tasks increases, it may loose the elasticity in consolidating the overlapping parameters, and as such show a steeper slope in the EWC plot of Figure \ref{fig: satur}.  

\section{Conclusions}
We presented a new approach in addressing the continual learning problem with deep neural networks. It is inspired by the randomization and compression techniques typically used in statistical analysis. We combined a simple importance sampling technique - leverage score sampling with the frequent directions concept and developed an online effective forgetting or compression mechanism that preserves meaningful information from previous and current task, enabling continual learning across a sequence of tasks. Despite its simple structure, the results on classification benchmark experiments (designed for the catastrophic forgetting issue) demonstrate its effectiveness as compared to recent state of the art. 

\end{document}